\documentclass[final]{article}

\PassOptionsToPackage{numbers, compress}{natbib}

\usepackage{neurips_2025}
\usepackage{amsmath}

\usepackage[utf8]{inputenc} 
\usepackage[T1]{fontenc}    
\usepackage{hyperref}       
\usepackage{url}            
\usepackage{booktabs}       
\usepackage{amsfonts}       
\usepackage{nicefrac}       
\usepackage{microtype}      
\usepackage{xcolor}         
\usepackage{algorithm}
\usepackage{algorithmic}
\usepackage{multirow}
\usepackage{graphicx}
\usepackage{svg}
\usepackage{amsthm}
\usepackage{tikz}
\usepackage{pgfplots}
\pgfplotsset{compat=1.18}

\usepackage{subcaption}  
\usepackage{xcolor}  
\usetikzlibrary{calc,fit}

\newcommand{\lineWithBackground}[1]{%
  \begin{tikzpicture}[baseline=-.5ex]
    \filldraw[fill=#1!15, draw=none, rounded corners=1pt] (-0.25,-0.12) rectangle (0.25,0.12);
    \draw[#1, line width=0.8pt] (-0.25,0) -- (0.25,0);
  \end{tikzpicture}%
}

\title{Dynamic Action Interpolation: A Universal Approach for Accelerating Reinforcement Learning with Expert Guidance}

\author{Wenjun Cao \\
  Independent Researcher \\
  wenjun.cao.research@gmail.com}

\begin{document}

\maketitle

\begin{abstract}
Reinforcement learning (RL) suffers from severe sample inefficiency, especially during early training, requiring extensive environmental interactions to perform competently. Existing methods tend to solve this by incorporating prior knowledge, but introduce significant architectural and implementation complexity. We propose Dynamic Action Interpolation (DAI), a universal yet straightforward framework that interpolates expert and RL actions via a time-varying weight $\alpha(t)$, integrating into any Actor-Critic algorithm with just a few lines of code and without auxiliary networks or additional losses. Our theoretical analysis shows that DAI reshapes state visitation distributions to accelerate value function learning while preserving convergence guarantees. Empirical evaluations across MuJoCo continuous control tasks demonstrate that DAI improves early-stage performance by over 160\% on average and final performance by more than 50\%, with the Humanoid task showing a 4$\times$ improvement early on and a 2$\times$ gain at convergence. These results challenge the assumption that complex architectural modifications are necessary for sample-efficient reinforcement learning.
\end{abstract}

\section{Introduction}
Reinforcement learning (RL) has performed strongly in complex control domains such as locomotion, manipulation, and robotics. However, RL methods remain notoriously sample-inefficient, particularly during early training when policies must explore high-dimensional state-action spaces without guidance. In environments like humanoid locomotion or dexterous manipulation, agents often require millions of interactions before achieving competent performance, severely limiting their feasibility in practical applications.

Numerous methods have sought to incorporate prior knowledge into RL \cite{tirumala2022behavior}, primarily by intervening during policy optimization or training. Some approaches leverage demonstrations to guide learning via auxiliary imitation losses or prioritized sampling \cite{hester2018deep, pmlr-v80-kang18a, nair2018overcomingexplorationreinforcementlearning, paine2019making}. In contrast, others shape reward functions based on expert behaviors using manually designed or generative models \cite{10.1109/ICRA48506.2021.9561333}. A parallel trend focuses on offline-to-online hybrid learning frameworks that fine-tune RL agents from demonstration-derived datasets \cite{nair2020awac, NEURIPS2023_c44a0428}. While effective, these methods often introduce substantial architectural complexity, auxiliary objectives, or multistage training pipelines, entangling prior knowledge integration with core learning dynamics.

This persistent pattern raises a fundamental question: \textit{Can we find a way to harness prior knowledge in reinforcement learning that is both algorithmically simple and universally applicable across different domains and methods?}

Instead of modifying policy architectures or training objectives, we propose returning to first principles by intervening directly at the action execution level—where agent decisions immediately impact environmental interaction. Our approach, Dynamic Action Interpolation (DAI), interpolates expert and RL-generated actions in real-time using a time-varying weight $\alpha(t)$, without requiring any changes to the objective function, loss terms, or network structure. Unlike residual RL methods \cite{johannink2018residualreinforcementlearningrobot} that rely on fixed base controllers or dynamic weighting approaches \cite{liu2023blendingimitationreinforcementlearning, zhao2022adaptive} that modify training objectives, DAI operates exclusively at the execution level, preserving the original learning dynamics. Implementable in just a few lines of code within any Actor-Critic algorithm, this simple yet powerful intervention offers a more robust and efficient mechanism for improving sample efficiency in reinforcement learning.

Our theoretical analysis reveals how DAI shapes state distributions to accelerate value function learning while preserving convergence guarantees. Experiments on MuJoCo continuous control tasks demonstrate that DAI improves early-stage performance by over 160\% on average and ultimately surpasses expert policy performance by more than 100\%. In the complex Humanoid environment, DAI achieves more than 4× improvement in early training and 2× advantage at convergence.

This paper makes the following contributions:
\begin{enumerate}
    \item Introducing DAI, a simple and universal framework that directly interpolates expert and RL actions at the action execution level.
    \item Developing a novel theoretical framework showing how DAI reshapes state visitation distributions to accelerate learning while maintaining asymptotic convergence properties.
    \item Demonstrating that DAI integrates seamlessly into any Actor-Critic algorithm with minimal code changes, requiring no new loss terms or networks.
    \item Showing DAI's strong performance and robustness across multiple continuous control benchmarks, both in accelerating early learning and improving final performance.
\end{enumerate}

These results demonstrate that returning to algorithmic simplicity can yield superior performance. They challenge the trend toward elaborate architectures and suggest broader applicability for prior knowledge integration in reinforcement learning.

\section{Related Work}
\paragraph{Prior Knowledge in RL.}
While integrating prior knowledge into RL has been widely studied, several approaches introduce added complexity, which may not always result in proportional improvements in performance. For instance, early demonstration-based methods~\cite{hester2018deep, nair2018overcomingexplorationreinforcementlearning, rajeswaran2018learningcomplexdexterousmanipulation} required elaborate auxiliary losses or reward engineering. Approaches leveraging demonstrations through replay mechanisms or policy constraints~\cite{vecerik2018leveragingdemonstrationsdeepreinforcement, paine2019making, pmlr-v80-kang18a} necessitate careful dataset balancing. Recent methods bridging offline and online learning~\cite{nair2020awac, NEURIPS2023_c44a0428, lee2021offlinetoonlinereinforcementlearningbalanced, beeson2022improvingtd3bcrelaxedpolicy, zhao2022adaptive, zheng2022online} attempt to address distribution shift but rely on specialized architectures, complex replay strategies, or sequence modeling techniques. Additional approaches~\cite{singh2020cog, tirumala2022behavior, sun2023smart, song2023hybridrlusingoffline} explore various mechanisms from fine-tuning to behavior priors. Nevertheless, all share a common limitation: they significantly increase implementation complexity through specialized architectures, auxiliary objectives, or multi-stage training pipelines. In contrast, DAI integrates prior knowledge directly at the action execution level without modifying learning objectives or network architectures, achieving true algorithm agnosticism with minimal code changes while maintaining or exceeding the performance of more complex methods.

\paragraph{Hybrid Methods in RL.}
A promising line of research has addressed efficiency challenges by directly intervening in the action space. Residual RL methods~\cite{johannink2018residualreinforcementlearningrobot, alakuijala2021residualreinforcementlearningdemonstrations, Davchev_2022} combine fixed base controllers with learned components but lack transition mechanisms. Objective modification approaches~\cite{liu2023blendingimitationreinforcementlearning, zhao2022adaptive} integrate expert guidance through training objectives, requiring substantial architectural changes. Adaptive expert methods~\cite{liu2023guidelearnerimitationlearning, dey2022jointimitationreinforcementlearningframework} use complex decision logic to determine when to leverage guidance. Structured progression approaches~\cite{czarnecki2018mixmatchagentcurricula, NEURIPS2021_31839b03, pmlr-v9-ross10a} implement curricula with explicit training stages, while theoretical frameworks~\cite{sun2018truncatedhorizonpolicysearch, swamy2023minimaxoptimalonlineimitation, ren2024hybridinversereinforcementlearning} provide elegant but complex mathematical formulations. Despite their theoretical diversity, these methods share a common limitation—they inevitably trade simplicity for expressiveness, introducing complex architectures, auxiliary objectives, or elaborate optimization procedures. In contrast, DAI achieves superior results with a simpler linear interpolation mechanism implemented in just a few lines of code.

\paragraph{Simplicity in RL.}
DAI aligns with the growing recognition that algorithmic simplicity can yield strong performance in reinforcement learning. Empirical studies \cite{10.5555/3504035.3504427, tucker2018mirageactiondependentbaselinesreinforcement, engstrom2020implementationmattersdeeppolicy, andrychowicz2020mattersonpolicyreinforcementlearning, wu2019behavior} have consistently shown that implementation details often matter more than algorithmic innovations, and claimed theoretical advantages may stem from implementation biases rather than core algorithm design. The field has increasingly embraced a minimalist perspective, focusing on reducing complexity while maintaining strong performance \cite{fujimoto2021minimalist, peng2019advantage, kumar2020conservative}. DAI continues this trend by providing a direct, implementation-friendly approach that performs well without complex architectural modifications.

In summary, DAI distinguishes itself from existing methods through its exceptional combination of implementation simplicity, theoretical guarantees, and strong empirical performance. It requires just a few lines of code added to any Actor-Critic algorithm while providing both early-stage acceleration and asymptotic optimality. Its proven effectiveness across diverse environments completes this unique blend of simplicity, universality, and performance that positions DAI distinctly among existing approaches.

\section{Method and Mathematical Framework}

\paragraph{Preliminaries and Notation.}
We analyze DAI within the standard Markov Decision Process (MDP) framework $(S, A, P, R, \gamma)$, where $S$ is the state space, $A$ is the action space, $P$ is the transition probability function, $R$ is the reward function, and $\gamma \in [0, 1)$ is the discount factor.

For a policy $\pi$, we define the state-action value function as:
\begin{equation}
Q^{\pi}(s, a) = \mathbb{E}\left[\sum_{t=0}^{\infty} \gamma^t r_t|s_0 = s, a_0 = a, \pi\right]
\end{equation}

The optimal policy $\pi^*$ maximizes the expected return from all states. Dynamic Action Interpolation (DAI) aims to combine an expert policy $\pi_E$ and a reinforcement learning policy $\pi_{RL}$ to achieve both sample efficiency and asymptotic optimality.

\paragraph{Dynamic Action Interpolation and Weight Function.}
Dynamic Action Interpolation directly implements linear interpolation between expert and RL policy outputs. Formally, for any state $s \in S$, the interpolated action is defined as:
\begin{equation}
a^{\text{mix}}(s) = (1 - \alpha(t)) \cdot a_E(s) + \alpha(t) \cdot a_{RL}(s)
\end{equation}
where $a_E(s)$ is the expert policy output, $a_{RL}(s)$ is the RL policy output, and $\alpha(t) \in [0, 1]$ is a time-varying weight function controlling the interpolation proportion. This formulation is applicable to both continuous action spaces (direct action vectors) and discrete action spaces (logits or probability distributions).

For the weight function $\alpha(t)$, we define a generalized framework:
\begin{equation}
\alpha(t) = \phi\left(\frac{t}{T_{change}}\right)
\end{equation}

where $\phi : [0, \infty) \rightarrow [0, 1]$ is any monotonically non-decreasing function satisfying $\phi(0) = 0$ and $\lim_{x\rightarrow\infty} \phi(x) = 1$.

In our practical implementation, we use the linear annealing function for its simplicity:
\begin{equation}
\alpha(t) = \min\left(\max\left(\frac{t}{T_{change}}, 0\right), 1\right)
\end{equation}

\paragraph{Algorithm Implementation.}
The DAI algorithm can be implemented with minimal code changes to existing actor-critic methods:

\begin{algorithm}[H]
\caption{Dynamic Action Interpolation (DAI)}
\begin{algorithmic}[1]
\REQUIRE Expert policy $\pi_E$, RL policy $\pi_{RL}$, environment \texttt{Env}, weight function $\alpha(t)$, total steps $T$
\FOR{$t = 1$ to $T$}
    \STATE Get current state $s$
    \STATE Compute interpolation weight $\alpha_t = \alpha(t)$
    \STATE Get expert action $a_E = \pi_E(s)$
    \STATE Get RL action $a_{RL} = \pi_{RL}(s)$
    \STATE Compute interpolated action $a^{\text{mix}} = (1 - \alpha_t) \cdot a_E + \alpha_t \cdot a_{RL}$
    \STATE Execute $a^{\text{mix}}$ in \texttt{Env}, observe reward $r$ and next state $s'$
    \STATE Update $\pi_{RL}$ using its learning algorithm
    \IF{episode ends}
        \STATE Reset environment
    \ENDIF
\ENDFOR
\end{algorithmic}
\end{algorithm}

\section{Theoretical Framework of DAI for General Actor-Critic Methods}

We have developed a theoretical framework for Dynamic Action Interpolation (DAI) that applies to general actor-critic methods independent of specific algorithm implementations. This section establishes how DAI's simple mechanism fundamentally transforms learning dynamics while maintaining theoretical guarantees.

\paragraph{Setup.}
We consider a standard Markov Decision Process (MDP) as introduced in Section 3. In the actor-critic framework, we have:
\begin{itemize}
    \item An actor policy $\pi_\theta(a|s)$ parameterized by $\theta$, which maps states to (possibly stochastic) actions
    \item A critic that estimates the state-action value function $Q^\pi(s, a)$ or state value function $V^\pi(s)$ to guide policy updates
    \item An expert policy $\pi_E$ providing demonstration behavior
\end{itemize}

The reinforcement learning objective is to find a policy $\pi^*$ that maximizes the expected return from all states. Actor-critic methods iteratively improve the policy $\pi_\theta$ based on value estimates from the critic.

\paragraph{Action Interpolation Mechanism.}
To integrate expert guidance without modifying the learning objective or the policy parameterization, DAI introduces an \textbf{action interpolation mechanism}:

\begin{equation}
a(s,t) = (1 - \alpha(t)) \cdot a_E(s) + \alpha(t) \cdot a_{RL}(s)
\end{equation}

where $a_E(s)$ and $a_{RL}(s)$ are the outputs of the expert and RL policies, respectively. The interpolation coefficient $\alpha(t) \in [0,1]$ is a monotonic schedule with $\alpha(0) = 0$ and $\lim_{t\to\infty}\alpha(t) = 1$.

This defines a deterministic action distribution:

\begin{equation}
p(a | s, t) = \delta(a = a(s,t)) = \delta(a = (1-\alpha(t)) \cdot a_E(s) + \alpha(t) \cdot a_{RL}(s))
\end{equation}

where $\delta(\cdot)$ is the Dirac delta function, representing a deterministic selection of the interpolated action.

Significantly, this mechanism directly modifies the action executed in the environment, but \textbf{does not constitute a separate policy}. The underlying RL policy $\pi_\theta$ continues to be trained using its original gradient objective.

\paragraph{State Distribution Shaping.}
By modifying the executed action during environment interaction, the interpolation mechanism alters the induced state visitation distribution. For any policy $\pi$, the discounted state visitation distribution is defined as:

\begin{equation}
d^\pi(s) = (1-\gamma)\sum_{t=0}^{\infty}\gamma^t P(s_t=s|\pi)
\end{equation}

Let $d^a(s)$ denote the state distribution induced by the interpolated actions. While the exact relationship between $d^a(s)$ and the component policies is complex due to the nonlinear dynamics of MDPs, we can characterize it with the following approximation under the assumption of local Lipschitz transition dynamics:

\begin{equation}
d^a(s) \approx (1-\alpha(t)) \cdot d^{\pi_E}(s) + \alpha(t) \cdot d^{\pi_\theta}(s)
\end{equation}

This approximation captures the intuition that as $\alpha(t)$ smoothly transitions from 0 to 1, the state distribution similarly gradually shifts from expert-like to RL-like.

For states with high value (denoted as $S_V = \{s \in S|V^*(s) > V_{\text{threshold}}\}$ for some threshold), we can posit that early in training:

\begin{equation}
P(s \in S_V|a(s,t)) > P(s \in S_V|\pi_{\theta,\text{initial}})
\end{equation}

This inequality holds when the expert policy effectively guides the agent to valuable regions of the state space. It explains why DAI can significantly accelerate learning in complex environments where exploration is challenging.

\paragraph{Value Learning Acceleration Hypothesis.}
Given that critic updates are based on samples drawn from the current behavior, the quality of the state distribution directly impacts value approximation. With the interpolated execution mechanism, the agent collects data from more informative regions:

\begin{equation}
\mathbb{E}_{s \sim d^a}[(V^{\pi_\theta}(s) - \hat{V}(s))^2] < \mathbb{E}_{s \sim d^{\pi_{\theta,\text{initial}}}}[(V^{\pi_\theta}(s) - \hat{V}(s))^2]
\end{equation}

Here, $\hat{V}(s)$ denotes the learned critic approximation of $V^{\pi_\theta}(s)$ under the current policy $\pi_\theta$, not of the optimal value function $V^*$. This hypothesis formalizes the intuition that by guiding exploration toward high-value regions, DAI improves the quality of samples used for critic training, resulting in faster convergence of value estimates. Importantly, this mechanism applies to any actor-critic algorithm, whether on-policy or off-policy, as it concerns the distribution of training data rather than the specific update rules.

\paragraph{Interactive Data Collection Benefits.}
DAI improves data efficiency and learning robustness by generating actions through real-time interpolation between expert and learned policies. These interpolated actions are computed online and conditioned on the current state, resulting in adaptive behavior that evolves with training progress. This mechanism enables the agent to leverage expert knowledge while actively exploring its environment, potentially generalizing beyond fixed demonstrations. Importantly, since the interpolated actions are executed in the environment and reinforced through immediate feedback, DAI maintains a closed-loop interaction where learning remains tightly coupled with actual behavior.

\paragraph{Transition from Imitation to Autonomy.}
As $\alpha(t)$ increases, the executed actions transition from being dominated by the expert to being fully determined by $\pi_\theta$. This induces a continuous curriculum of interaction dynamics without explicitly modifying the learning target. The state distribution over which gradients are computed evolves as:

\begin{equation}
\nabla_\theta J(\theta) = \mathbb{E}_{s \sim d^a}\left[\mathbb{E}_{a \sim \pi_\theta(\cdot|s)}[\nabla_\theta \log \pi_\theta(a|s)Q^{\pi_\theta}(s,a)]\right]
\end{equation}

While the policy gradient formula is unchanged, the state distribution $d^a$ over which the expectation is taken evolves with $\alpha(t)$. This creates a curriculum effect, where the agent initially learns from states visited under expert guidance and gradually shifts to states resulting from its decision-making.

This natural progression mimics human learning, where initial guidance leads to autonomous skill development. The gradual nature of this transition allows the agent to build upon expert knowledge while exploring beyond it, potentially surpassing the expert's capabilities.

\paragraph{Theoretical Analysis of Asymptotic Performance.}
We now show that DAI preserves the long-term performance of any actor-critic algorithm that converges to a stationary policy. While DAI modifies the actions used during training, it does not alter the underlying learning objective or parameter updates. The key question is whether interpolating with expert actions impairs the policy's ability to reach its optimal or stable performance in the long run.

We assume that the underlying actor-critic method induces a sequence of policies $\{\pi_\theta^{(t)}\}$ which converges to a stationary policy $\pi_{\theta,\text{final}}$ in the limit as training progresses. This assumption does not require convergence of the learning algorithm in a strict optimization sense but only that the behavior policy stabilizes over time.

Under this assumption, and given that the action interpolation coefficient $\alpha(t)$ satisfies $\lim_{t \to \infty} \alpha(t) = 1$, the executed action $a(s,t)$ deterministically converges to $a_{RL}(s)$ as $\alpha(t) \to 1$. As a result, the long-term behavior of the DAI-enhanced agent converges to that of the base policy $\pi_{\theta,\text{final}}$, and we have:

\begin{equation}
\lim_{t\to\infty} \mathbb{E}[G_0 | a(s,t)] = \mathbb{E}[G_0 | \pi_{\theta,\text{final}}]
\end{equation}

This guarantees that DAI does not compromise asymptotic performance while still providing the early learning benefits of expert guidance. When the expert policy is suboptimal compared to what the RL algorithm could eventually learn, DAI's gradual transition to fully autonomous decision-making ensures the agent can discover solutions that potentially exceed the expert's capabilities.

\paragraph{Algorithm Independence.}
Since DAI operates exclusively in the action execution phase, it is compatible with any actor-critic algorithm, either on-policy or off-policy, without requiring changes to the underlying training objectives or architecture. Whether applied to various on-policy or off-policy methods, the core mechanism remains identical: interpolating between expert demonstrations and learned policy actions during environment interaction while maintaining the original optimization objective for policy learning.

This universality makes DAI an appealing meta-algorithm that can enhance a wide range of actor-critic methods without fundamentally redesigning their update rules or objectives. The theoretical benefits, including improved state distribution, accelerated value learning, and guaranteed convergence, apply broadly across the actor-critic family.

Through this unified theoretical framework, we establish that DAI's simple action interpolation mechanism generally offers significant advantages for actor-critic methods, accelerating early learning without compromising asymptotic performance. This analysis complements our empirical results, explaining why such a straightforward approach leads to substantial improvements across diverse environments.

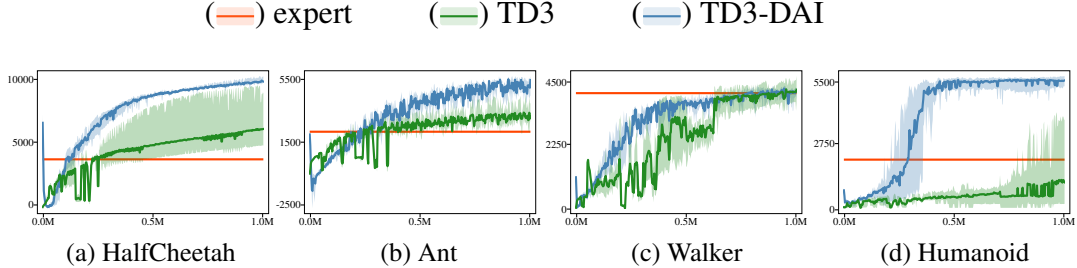
\begin{figure}[t]
    \centering
    \definecolor{orange}{RGB}{255,83,19}
    \definecolor{blue}{RGB}{70,130,180}
    \definecolor{green}{RGB}{34,139,34}
    
    \centering
    {\large
    ({\lineWithBackground{orange}}) expert \hspace{0.8cm}
    ({\lineWithBackground{green}}) TD3 \hspace{0.8cm}
    ({\lineWithBackground{blue}}) TD3-DAI \hspace{0.8cm}
    }
    
    \vspace{0.5cm}
    
    \hspace{-0.02\textwidth}
    \makebox[0.23\textwidth][c]{%
        \begin{tikzpicture}[inner sep=0pt, outer sep=0pt]
            \node[inner sep=0] (img) at (0,0) {\includegraphics[width=0.21\textwidth]{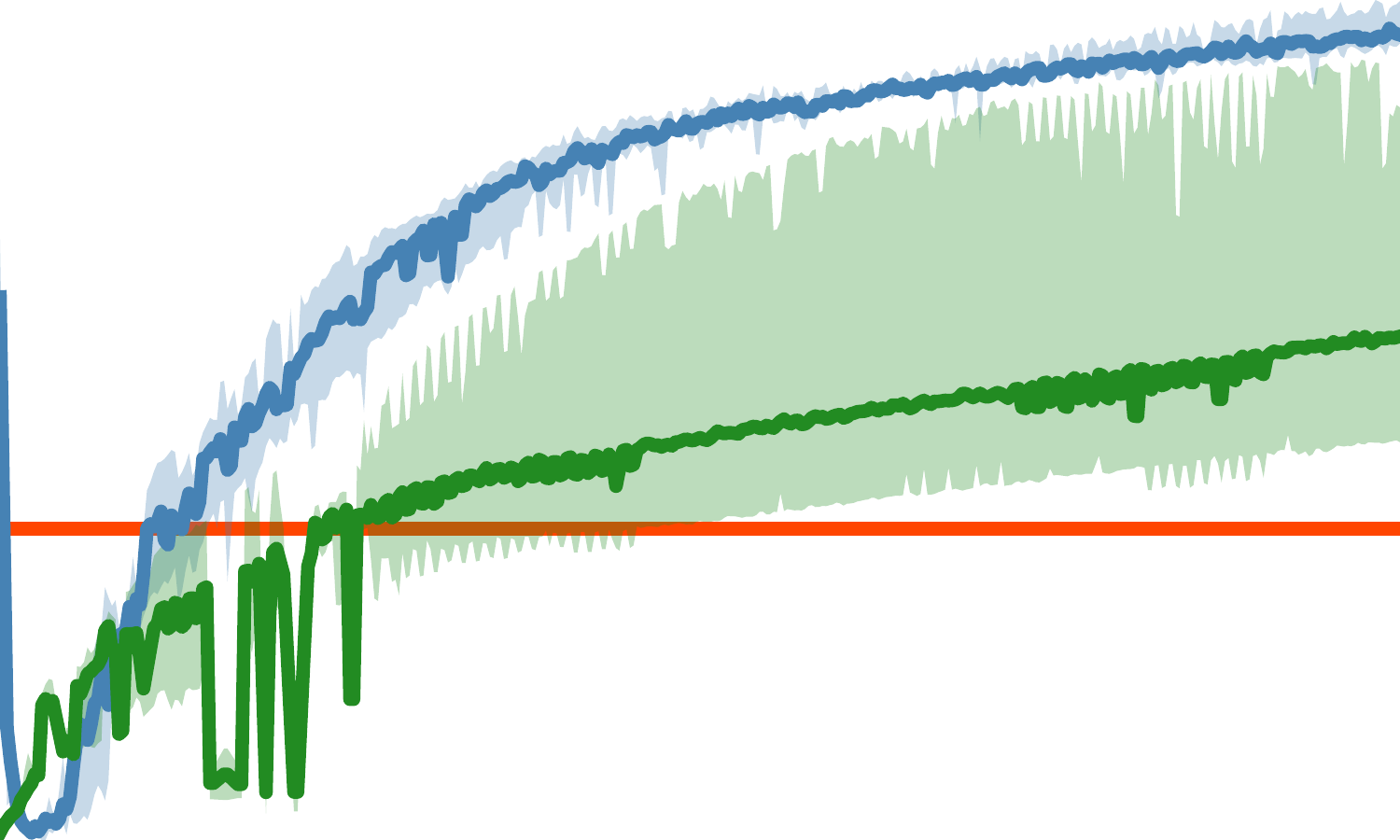}};
            
            \node[draw, very thin, fit=(img), inner sep=2pt] (frame) {};
            
            \draw ($(frame.south west)+(0.08,0)$) -- ++(0,-0.05) node[below, font=\fontsize{4}{5}\selectfont] {0.0M};
            \draw ($(frame.south west)!0.5!(frame.south east)$) -- ++(0,-0.05) node[below, font=\fontsize{4}{5}\selectfont] {0.5M};
            \draw ($(frame.south east)-(0.08,0)$) -- ++(0,-0.05) node[below, font=\fontsize{4}{5}\selectfont] {1.0M};
            
            \draw ($(frame.south west)+(0,0.115)$) -- ++(-0.05,0) node[left, font=\fontsize{4}{5}\selectfont] {0};
            \draw ($(frame.south west)!0.5!(frame.north west)$) -- ++(-0.05,0) node[left, font=\fontsize{4}{5}\selectfont] {5000};
            \draw ($(frame.north west)-(0,0.115)$) -- ++(-0.05,0) node[left, font=\fontsize{4}{5}\selectfont] {10000};
        \end{tikzpicture}%
    }%
    \hspace{0.025\textwidth}
    \makebox[0.23\textwidth][c]{%
        \begin{tikzpicture}[inner sep=0pt, outer sep=0pt]
            \node[inner sep=0] (img) at (0,0) {\includegraphics[width=0.21\textwidth]{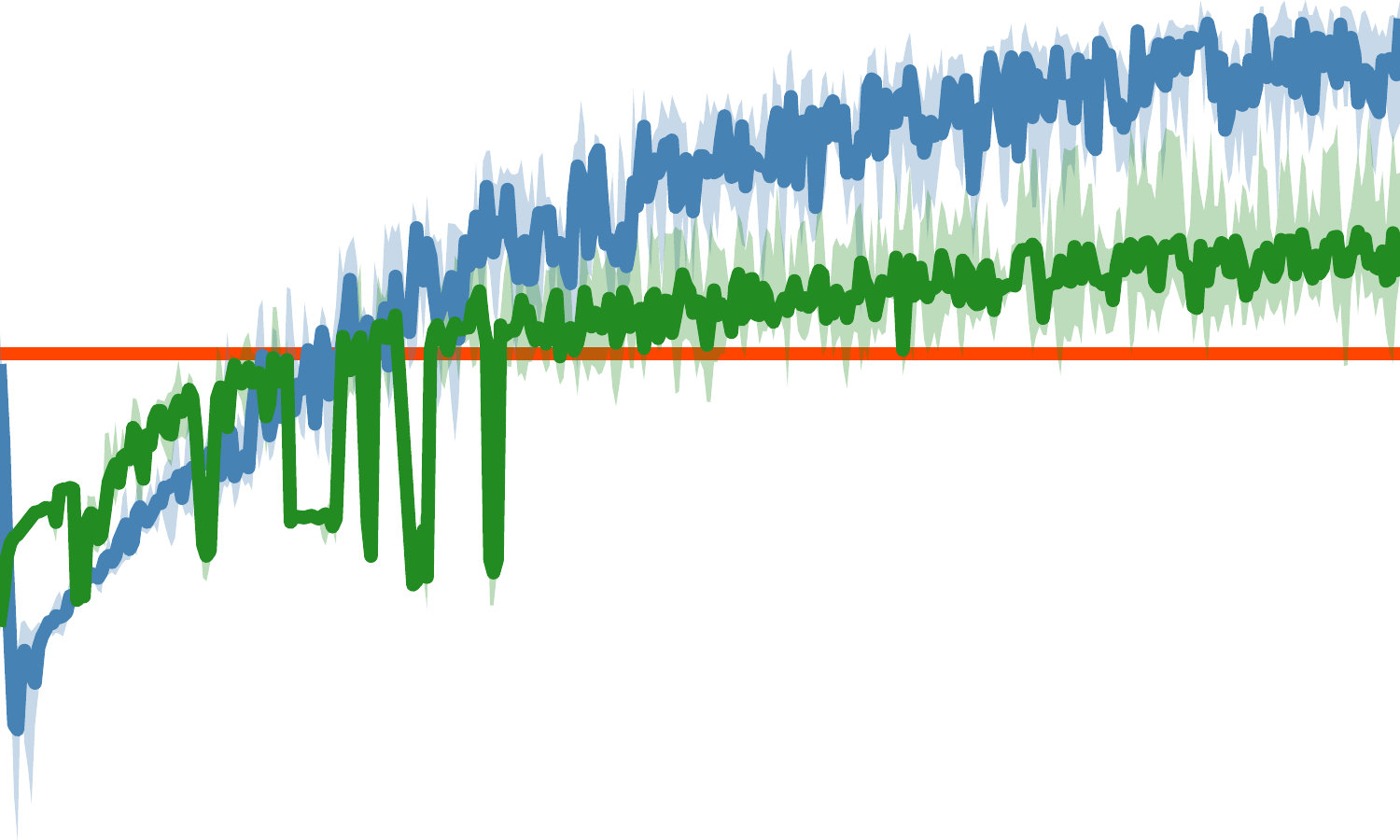}};
            \node[draw, very thin, fit=(img), inner sep=2pt] (frame) {};
            
            \draw ($(frame.south west)+(0.08,0)$) -- ++(0,-0.05) node[below, font=\fontsize{4}{5}\selectfont] {0.0M};
            \draw ($(frame.south west)!0.5!(frame.south east)$) -- ++(0,-0.05) node[below, font=\fontsize{4}{5}\selectfont] {0.5M};
            \draw ($(frame.south east)-(0.08,0)$) -- ++(0,-0.05) node[below, font=\fontsize{4}{5}\selectfont] {1.0M};
            
            \draw ($(frame.south west)+(0,0.115)$) -- ++(-0.05,0) node[left, font=\fontsize{4}{5}\selectfont] {-2500};
            \draw ($(frame.south west)!0.5!(frame.north west)$) -- ++(-0.05,0) node[left, font=\fontsize{4}{5}\selectfont] {1500};
            \draw ($(frame.north west)-(0,0.115)$) -- ++(-0.05,0) node[left, font=\fontsize{4}{5}\selectfont] {5500};
        \end{tikzpicture}%
    }%
    \hspace{0.025\textwidth}
    \makebox[0.23\textwidth][c]{%
        \begin{tikzpicture}[inner sep=0pt, outer sep=0pt]
            \node[inner sep=0] (img) at (0,0) {\includegraphics[width=0.21\textwidth]{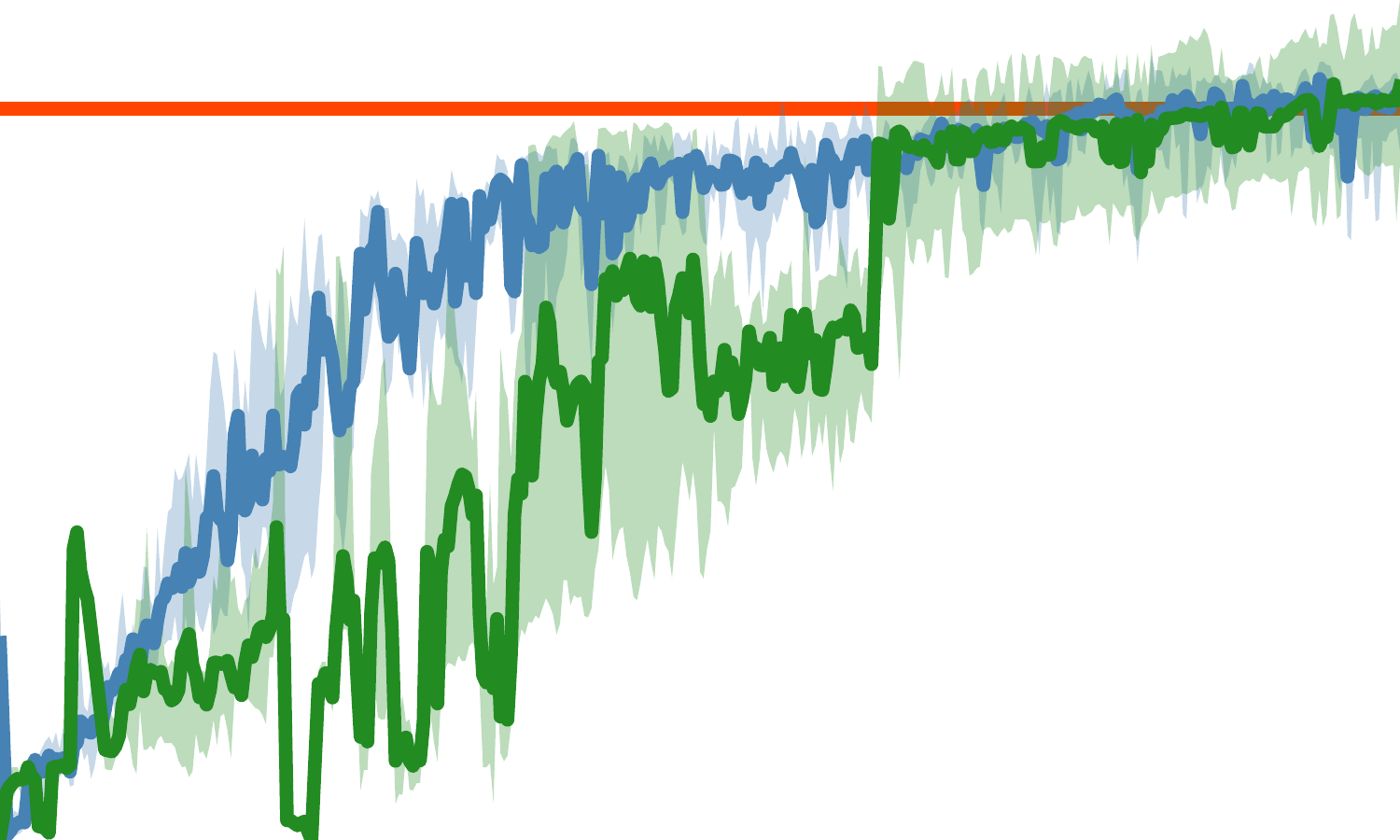}};
            \node[draw, very thin, fit=(img), inner sep=2pt] (frame) {};
            
            \draw ($(frame.south west)+(0.08,0)$) -- ++(0,-0.05) node[below, font=\fontsize{4}{5}\selectfont] {0.0M};
            \draw ($(frame.south west)!0.5!(frame.south east)$) -- ++(0,-0.05) node[below, font=\fontsize{4}{5}\selectfont] {0.5M};
            \draw ($(frame.south east)-(0.08,0)$) -- ++(0,-0.05) node[below, font=\fontsize{4}{5}\selectfont] {1.0M};
            
            \draw ($(frame.south west)+(0,0.06)$) -- ++(-0.05,0) node[left, font=\fontsize{4}{5}\selectfont] {0};
            \draw ($(frame.south west)!0.485!(frame.north west)$) -- ++(-0.05,0) node[left, font=\fontsize{4}{5}\selectfont] {2250};
            \draw ($(frame.north west)-(0,0.15)$) -- ++(-0.05,0) node[left, font=\fontsize{4}{5}\selectfont] {4500};
        \end{tikzpicture}%
    }%
    \hspace{0.025\textwidth}
    \makebox[0.23\textwidth][c]{%
        \begin{tikzpicture}[inner sep=0pt, outer sep=0pt]
            \node[inner sep=0] (img) at (0,0) {\includegraphics[width=0.21\textwidth]{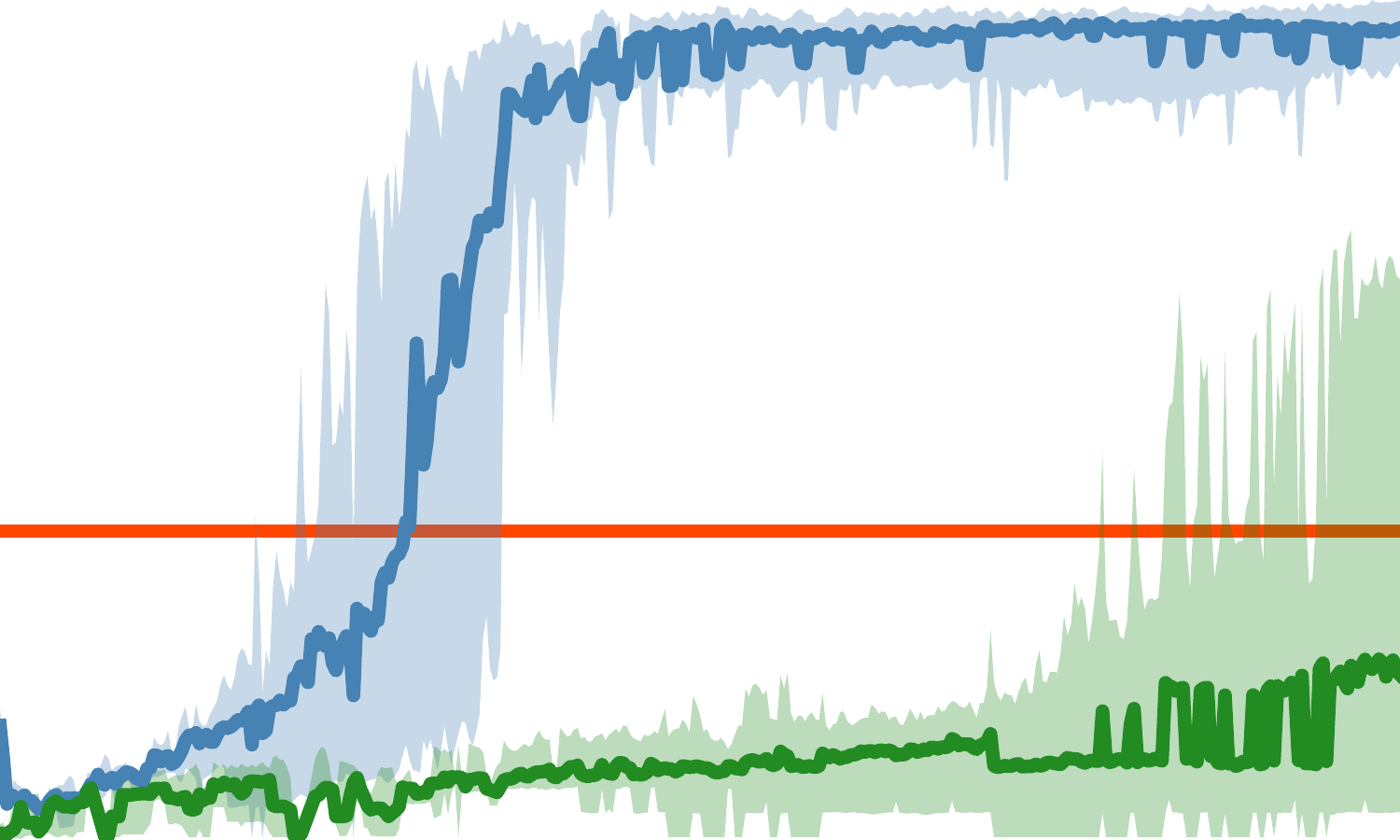}};
            \node[draw, very thin, fit=(img), inner sep=2pt] (frame) {};
            
            \draw ($(frame.south west)+(0.08,0)$) -- ++(0,-0.05) node[below, font=\fontsize{4}{5}\selectfont] {0.0M};
            \draw ($(frame.south west)!0.5!(frame.south east)$) -- ++(0,-0.05) node[below, font=\fontsize{4}{5}\selectfont] {0.5M};
            \draw ($(frame.south east)-(0.08,0)$) -- ++(0,-0.05) node[below, font=\fontsize{4}{5}\selectfont] {1.0M};
            
            \draw ($(frame.south west)+(0,0.05)$) -- ++(-0.05,0) node[left, font=\fontsize{4}{5}\selectfont] {0};
            \draw ($(frame.south west)!0.49!(frame.north west)$) -- ++(-0.05,0) node[left, font=\fontsize{4}{5}\selectfont] {2750};
            \draw ($(frame.north west)-(0,0.15)$) -- ++(-0.05,0) node[left, font=\fontsize{4}{5}\selectfont] {5500};
        \end{tikzpicture}%
    }
    
    \par\vspace{0.1cm}
    \makebox[0.225\textwidth][c]{(a) HalfCheetah}%
    \hspace{0.03\textwidth}%
    \makebox[0.225\textwidth][c]{(b) Ant}%
    \hspace{0.03\textwidth}%
    \makebox[0.225\textwidth][c]{(c) Walker}%
    \hspace{0.03\textwidth}%
    \makebox[0.225\textwidth][c]{(d) Humanoid}

    \caption{Performance comparison of TD3-DAI against TD3 and expert policies across four MuJoCo environments. TD3-DAI significantly outperforms both baselines in all environments. In the early training phase, TD3-DAI substantially improves over TD3, with the most pronounced gains in the Humanoid environment. This advantage persists throughout the training process. Across all environments, TD3-DAI ultimately surpasses TD3 and expert policies by a considerable margin. Data points represent median performance across 6 independent runs with outliers removed using the IQR method. Shaded regions indicate 95\% confidence intervals computed via bootstrapping. Orange horizontal lines represent expert policy performance benchmarks.}
    \label{fig:performance_comparison}
\end{figure}

\section{Experiments}

\subsection{Experimental Goals and Setup}
The primary goal of our experimental evaluation is to assess how effectively DAI accelerates early learning compared to standard reinforcement learning methods, while also evaluating its ability to outperform expert policies eventually. Additionally, we aim to verify the consistency of DAI's performance across environments with varying complexity. To achieve this, we compare TD3-DAI against standard TD3 and expert baselines on four widely used MuJoCo continuous control environments from OpenAI Gym: HalfCheetah-v5, Ant-v5, Walker2d-v5, and Humanoid-v5. These environments span a range of increasing dimensionality and control complexity, providing a comprehensive testbed for DAI's performance.

For all experiments, we compare three distinct approaches: TD3-DAI, our proposed method that incorporates Dynamic Action Interpolation into TD3 \cite{pmlr-v80-fujimoto18a}, blending expert actions and RL-generated actions with a time-varying weight function; TD3, the standard reinforcement learning baseline, which learns entirely from scratch without expert guidance, representing the conventional deep reinforcement learning approach; and expert, a baseline derived from behavior cloning (BC) \cite{10.5555/647636.733043}, directly trained on demonstration data, embodying pure imitation learning.

To ensure rigorous evaluation, we standardized several key experimental conditions. First, a reference model was trained using vanilla SAC \cite{haarnoja2018softactorcriticoffpolicymaximum} for 500,000 environment steps per environment. Transition data from 20 episodes of this trained model was then collected and used to train the final expert policy via behavior cloning (BC). Both TD3-DAI and TD3 were evaluated using identical hyperparameters, including learning rates, network architectures, and batch sizes, to isolate the effect of the action fusion mechanism. Each algorithm configuration was assessed across 6 independent runs, with performance reported as the median and 95\% confidence intervals (shaded regions) computed via bootstrapping.

\begin{table}[t]
\centering
\begin{tabular}{lccc}
\toprule
Environment  & TD3 & TD3+DAI \\
\midrule
HalfCheetah-v5 &  3375.0$\pm$1425.8 & 6306.3$\pm$754.2 \\
Ant-v5 &  1929.9$\pm$451.4 & 2895.5$\pm$348.0 \\
Walker2d-v5 & 1732.2$\pm$876.2 & 2985.5$\pm$683.2 \\
Humanoid-v5 & 348.7$\pm$213.6 & 1861.1$\pm$1531.6 \\
\bottomrule
\end{tabular}
\vspace{0.25cm}  
\caption{Early performance comparison after 0.25 million training steps. The results show mean reward$\pm$standard deviation across six independent runs for each environment.}
\label{tab:early_performance}
\end{table}

\subsection{Results and Analysis}

\paragraph{Early Learning Performance.}
As shown in Table~\ref{tab:early_performance}, during early training (0.25M steps), TD3-DAI achieves an average reward improvement of 160.5\% compared to standard TD3 across all environments. This substantial early-stage acceleration confirms the primary benefit of integrating expert guidance with RL. The improvement is most significant in the Humanoid environment, where TD3-DAI demonstrates a 434\% performance increase, highlighting the particular effectiveness of DAI in high-dimensional state-action spaces where exploration is challenging.

\paragraph{Final Performance Comparison.}
Table~\ref{tab:final_performance} presents the final performance after 1M training steps. TD3-DAI maintains its advantage, achieving an average reward improvement of 52.8\% over standard TD3 and 101.2\% over the expert policy. This confirms that DAI accelerates early learning and enables the discovery of policies that ultimately surpass both the guiding expert and conventional RL methods. In the Humanoid environment, TD3-DAI achieves a 2.2× performance advantage over TD3, demonstrating DAI's ability to leverage expert guidance for enhanced asymptotic performance effectively.

\paragraph{Learning Dynamics.}
Figure~\ref{fig:performance_comparison} illustrates the complete learning trajectories across all environments. The curves reveal that TD3-DAI learns faster initially and maintains a more stable learning progress throughout training. The shaded confidence intervals demonstrate reduced variance in TD3-DAI's performance compared to standard TD3, particularly in the early stages. This stability is consistent with our theoretical analysis, as the expert guidance effectively constrains exploration to high-value regions of the state-action space.

\paragraph{Correlation with Environment Complexity.} The extent of improvement achieved by DAI is strongly correlated with the complexity of the environment. In high-dimensional environments like Humanoid, the performance gap between DAI and the baseline methods is most pronounced during both the early and final stages of training. This observation supports our theoretical analysis in Section 4, where we discussed how DAI reshapes the state visitation distribution. In environments with challenging exploration dynamics, expert demonstrations guide the agent towards high-value regions of the state space, accelerating early learning. As training progresses, the reinforcement learning component gradually takes over, optimizing beyond the expert’s capabilities and refining the policy further. This process highlights DAI's ability to both accelerate initial learning and enable long-term improvements through autonomous exploration.

\begin{table}[t]
\centering
\begin{tabular}{lccc}
\toprule
Environment & expert & TD3 & TD3+DAI \\
\midrule
HalfCheetah-v5 & 3442.8$\pm$4183.3 & 6653.2$\pm$2317.9 & 9906.3$\pm$250.5 \\
Ant-v5 & 2248.6$\pm$898.6 & 3419.6$\pm$872.6 & 4520.5$\pm$1638.9 \\
Walker2d-v5 & 4031.8$\pm$164.2 & 3780.5$\pm$1003.7 & 4079.6$\pm$120.0 \\
Humanoid-v5 & 2163.7$\pm$1134.6 & 2088.8$\pm$2197.3 & 4646.2$\pm$2250.0 \\
\bottomrule
\end{tabular}
\vspace{0.25cm}  
\caption{Final performance comparison after 1 million training steps. The results show mean reward$\pm$standard deviation across six independent runs for each environment.}
\label{tab:final_performance}
\end{table}

\section{Discussion}
This work demonstrates that substantial improvements in reinforcement learning efficiency can be achieved through a simple yet fundamental intervention: interpolating expert and learned actions at the execution level without modifying network architectures, objectives, or optimization procedures. Dynamic Action Interpolation (DAI) offers a universally applicable mechanism that accelerates value learning by shaping state visitation distributions, achieving consistent gains across continuous control benchmarks, particularly in high-dimensional environments. Our results challenge the assumption that effective integration of prior knowledge requires architectural complexity and suggest that revisiting core principles of policy execution can unlock new opportunities for scalable and practical RL. Future work could explore adaptive interpolation schedules, multi-source expert integration, and broader applications beyond continuous control.

\bibliographystyle{unsrtnat}  
\bibliography{references}      


\end{document}